\documentclass[10pt,twocolumn,letterpaper]{article}

\usepackage{3dv}
\usepackage{times}
\usepackage{epsfig}
\usepackage{graphicx}
\usepackage{amsmath}
\usepackage{amssymb}
\usepackage{cuted}
\usepackage{caption}

\DeclareRobustCommand*{\ora}{\overrightarrow}

\usepackage[pagebackref=true,breaklinks=true,letterpaper=true,colorlinks,bookmarks=false]{hyperref}

\threedvfinalcopy 

\def\threedvPaperID{****} 

\ifthreedvfinal\fi
\begin{document}

\title{Interactive Sketching of Mannequin Poses}

\author{Gizem Unlu\\
University College London\\
{\tt\small g.unlu@cs.ucl.ac.uk}
\and
Mohamed Sayed\\
University College London\\
{\tt\small m.sayed@cs.ucl.ac.uk}
\and
Gabriel Brostow\\
University College London\\
{\tt\small g.brostow@cs.ucl.ac.uk}
}

\maketitle
\thispagestyle{empty}
\begin{strip}
\centering
    \vspace{-2\baselineskip}
        \includegraphics[width=1.0\linewidth]{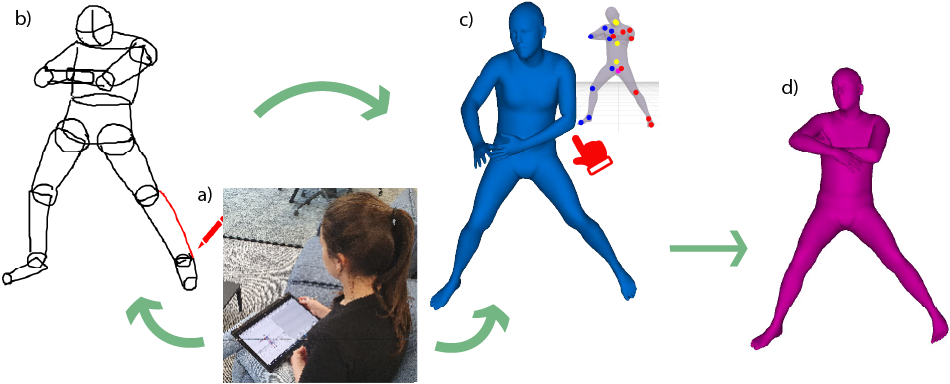}
    \vspace{-1.7\baselineskip}
        \captionof{figure}{Overview of our interactive sketch-based 3D figure posing. From figure sketches (b) we infer an initial 3D prediction (c). The user (a) then iterates on the sketch, or on the 3D pose using Forward and Inverse Kinematics handles to achieve a final refined pose (d).}
    \vspace{-\baselineskip}
    \label{fig:teaser}
\end{strip}
\begin{abstract}
\vspace{-1\baselineskip}
   It can be easy and even fun to sketch humans in different poses. In contrast, creating those same poses on a 3D graphics ``mannequin'' is comparatively tedious. Yet 3D body poses are necessary for various downstream applications. We seek to preserve the convenience of 2D sketching while giving users of different skill levels the flexibility to accurately and more quickly pose\slash refine a 3D mannequin. 
   
At the core of the interactive system, we propose a  machine-learning model for inferring the 3D pose of a CG mannequin from sketches of humans drawn in a cylinder-person style. Training such a model is challenging because of artist variability, a lack of sketch training data with corresponding ground truth 3D poses, and the high dimensionality of human pose-space. Our unique approach to synthesizing vector graphics training data underpins our integrated ML-and-kinematics system. We validate the system by tightly coupling it with a user interface, and by performing a user study, in addition to quantitative comparisons. 
\end{abstract}
\vspace{-2\baselineskip}
\section{Introduction}
\let\thefootnote\relax\footnotetext{\href{http://visual.cs.ucl.ac.uk/pubs/sketch2mannequin/index.html}{http://visual.cs.ucl.ac.uk/pubs/sketch2mannequin}}
Sketching people's body poses is challenging but fun. The artist's aim could be just creative pleasure, \eg there are hundreds of YouTube videos showcasing body-pose drawing. Or the artist may have some specific downstream task such as story-boarding an action film or drafting a comic book. Sketching humans, as in \cite{Lee1984marvel}, is an often-recommended strategy for artists to get started. Assemblies of primitives help the artist to figure out framing and pose, before embellishing with clothes and facial features.

We seek to give artists who focus on people's poses the fun and convenience of sketching, while iterating to end up with usable and editable 3D human mannequin geometry. On one hand, we are inspired by sites like \href{https://figurosity.com/figure-drawing-poses}{Figurosity} that stretch skilled artists so they can hand-draw realistically proportioned people in interesting poses. On the other hand, we wish to emulate the accessibility and 2D-to-3D functionality of Teddy~\cite{Igarashi1999siggraph} and MonsterMash~\cite{Dvoroznak2020tog}.

To make progress and lower the barrier to entry, we propose a hybrid machine learning (ML) and kinematics system. We want the user to manipulate the sketch or the pose of a human mannequin, while keeping the body shape fixed. 

\textbf{Overview} Through our web interface, a user sketches on a blank canvas as pictured in Fig~\ref{fig:teaser}.b. The sketch should depict one person in the desired pose. Their body should be made up of cylinders, an ellipsoid for the head, and with circles at the joints. The sketch can be imperfect, with some parts missing, with moderately messy lines, and out of proportion limbs. Then our neural-network based model interprets the sketch as a posed 3D mannequin, as shown in Fig~\ref{fig:teaser}.c. The user can explore other poses by redrawing sections of the sketch. Equally, our generated mannequin has basic rigging and joint limits that support the user in obtaining a refined pose  (Fig~\ref{fig:teaser}.d) through both forward kinematics (FK) and inverse kinematics (IK) interaction handles.  

To our knowledge, this is the first human-in-the-loop system for making and editing mannequin poses based on such sketch inputs. Our main technical contribution is a vector-graphics data synthesis and augmentation algorithm designed specifically to a) overcome the absence of real paired training data for sketches with 3D, and to b) cope with the highly variable and sometimes unrealistic sketching styles of beginners. 

\section{Related Work}
Here, we recap examples of sketch-based figure modeling, ranging from general-purpose sketching to our nearest neighbors in articulated creature sketching.

\noindent\textbf{Classic 3D from Sketches:} Starting with Teddy~\cite{Igarashi1999siggraph}, a user could learn to slightly adjust their drawing style to hint to the system what 3D shape was desired. FiberMesh~\cite{Nealen2007siggraph} builds on Teddy's interactive sketch modeling and blobby-surface optimization, to enable more controlled modeling. Some sketch lines serve as control curves and remain on the constructed 3D surface, where the user can push and pull them. This ability to refine the output is important to us.

In contrast to Teddy and FiberMesh,  ILoveSketch~\cite{Bae2008uist} and EverybodyLovesSketch~\cite{Bae2009uist} allow the artist to determine construction surfaces by sketching and drawing 2D curves on planes in multiple views. This works well for fairly skilled artists and man-made shapes. Between Teddy and ILoveSketch, \cite{Rivers2010siggraph} is multi-view, silhouette-based, and includes Boolean operations, where the 2D sketches must capture the shape from two orthographic views, \eg the top and side.

In this space, \cite{Kraevoy2009eurog, Gingold2009siggrapha} and \cite{Olsen2011cga} sit closest to our use-cases. 
\cite{Kraevoy2009eurog} expects a template 3D model \eg of a dog.  
Their template is deformed to align to the sketch contour via HMM feature correspondences between sketch strokes and template vertices. 
Similarly, \cite{Gingold2009siggrapha} presents an easy-to-use single view method, but requires descriptive stroke annotations \eg normals and length information, and object part annotations of primitive shapes. 
In contrast, we ask our users to draw primitives without needing annotations, and infer the SMPL parameters of a human-specific learned feature representation~\cite{Loper2015tog}. NaturaSketch~\cite{Olsen2011cga} uses the contour inflation technique in Teddy. 
Unlike ours, their interface does not allow the user to modify or refine the produced 3D shape. Another approach focuses on sketches of symmetric shapes such as animals in a relatively symmetric pose\cite{Entem2015cg}. The structurally symmetric parts are determined from the strokes in a contour image, and the depth hierarchy of these parts is determined automatically. Only contour sketches from a side-view can be constructed. Similarly, \cite{Dvoroznak2018sbim} is better at capturing fine details from the sketch, but requires more input than \cite{Entem2015cg}, namely a depth hierarchy of the semantic parts in the sketch. 

\begin{figure*}[ht]
  \includegraphics[width=\textwidth]{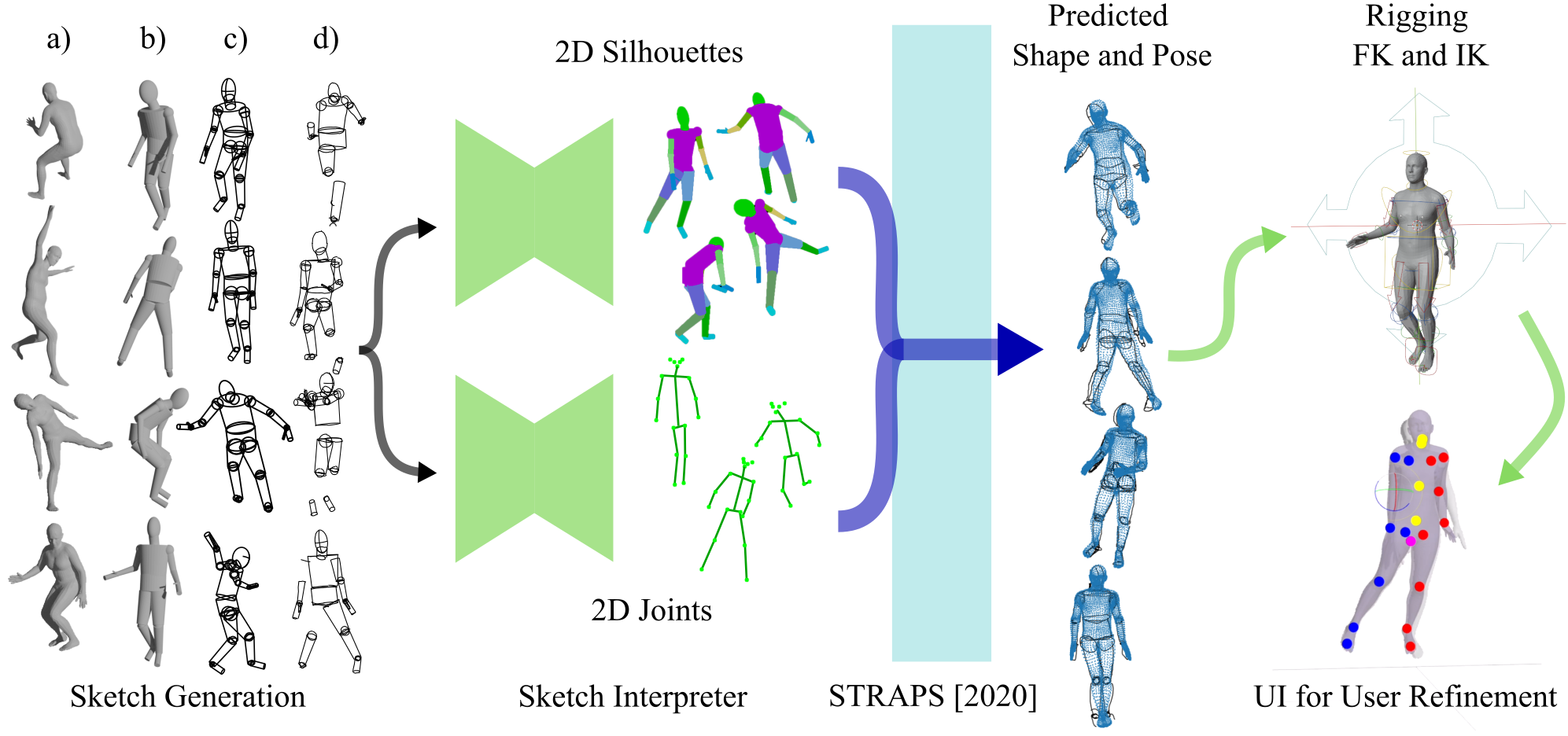}
  \vspace{-2\baselineskip}
  \caption{Overview of our sketch-based mannequin poser. (a) We sample SMPL poses and (b) generate our novel 3D primitive human model by placing 3D primitives on each sampled pose. (c) To generate sketches, we render the 3D primitive human to a `perfect' clean sketch. (d) Our augmentation scheme alters the clean sketches to mimic human-made sketches during network training. Given a figure sketch, our \emph{Sketch Interpreter} predicts 2D labeled silhouettes and joints from which 3D body pose and shape are inferred. The user can refine predicted poses or finetune their sketch interactively using our easy web-based user interface. Please note the poses in the figure are randomly selected samples and are unrelated through (a-d).}
  \vspace{-1\baselineskip}
    \label{fig:network}
\end{figure*} 
\noindent\textbf{Data-Driven Learning for Reconstruction:} Machine learning (ML) models are emerging that seek to capture correlations between sketches and their corresponding 3D shapes. 
Sketch2Pose~\cite{brodt2022sketch2pose} is concurrent work to our own, and despite being non-interactive, is our closest-neighbor, having an ML approach to finding at least the 2D joints, and using the same SMPL body representation~\cite{Loper2015tog}. We discuss \cite{brodt2022sketch2pose} further in the supplementary material. However, since realistic sketch-to-3D paired data is difficult to obtain in large enough quantities, many approaches exploit the ShapeNet~\cite{Chang2015arxiv} 3D repository, and seek to render its models in sketch-like Non-Photorealistic (NPR) styles, \eg Suggestive~\cite{DeCarlo2003siggraph} or Neural Contours~\cite{Liu2020cvpr}.

As with photo-based 3D inference of shapes~\cite{tang2021skeletonnet}, convolutional neural networks (CNNs) are being explored to handle sketches~\cite{Lun20173dv, Wang2020arxiv, Liu2019arxiv, smirnov2021patches, Li20133dor}. Given multiple sketches from different views, the network in \cite{Lun20173dv} infers depth and normal maps to construct 3D point clouds.
While the results on these are remarkable, the network requires carefully rendered orthographic drawings from a side or frontal view. It is not suitable for amateur sketches. 

In contrast, \cite{Wang2020arxiv} takes on the task of 3D point cloud reconstruction from a single sketch. Their model is composed of an off-the-shelf sketch synthesis network~\cite{Liu2019arxiv}, a sketch standardization module, and a reconstruction network. For generalization, augmentation includes deformations such as distortion, dilation, and erosion. They compare to previous works that use edge maps or NPR algorithms for sketch dataset generation, and show that training with their sketches generalizes better to real human drawings. That approach is scrutinized among many others in the large-scale study of Yue~\etal~\cite{Yue20203dv}. Yue~\etal determine the key challenges of working with sketch data. The most prominent differences between sketch and RGB image inputs are the former's sparsity and variation in sketching styles, and the imperfect nature of free-hand sketches in terms of perspective. They conduct experiments with three objects classes from ShapeNet. They also show that training with multiple categories vs. a single category hurts performance. On the basis of their evaluations, we compare to SynDraw~\cite{wailly2019line} as a data-augmentation baseline. 


A leading work for sketch-based modeling is SketchCNN~\cite{Li2018tog}. They have a two-stage approach instead of directly inferring the 3D geometry from sketches. First, an input sketch is mapped to an intermediate flow field. This representation contains local curvature information from the sketched object. Then a second network predicts the depth and normal maps corresponding to the input sketch and inferred flow-field. The user needs to distinguish contour strokes from other strokes. The interactive user interface allows for sketching from multiple views and 3D refinement. Also, the system can handle additional hints about curvature and depth by stroke annotation. This is still less input than counterpart BendSketch\cite{Li2017tog} requires, \ie to separately annotate each stroke \eg ridge/valley, curvature line, depth discontinuity, and boundary.

Delanoy~\etal~\cite{Delanoy2018cgit} present a modeling tool for predicting volumetric occupancy grids from sketches. Their pipeline has an initial single-view volume prediction step utilizing a user-drawn sketch from a viewpoint. The user can continue refining the shape from different viewpoints, and updated volumes are obtained iteratively. Their approach is based on two-volume predicting networks: a single-view prediction CNN and an updater CNN. Initially, the single view CNN is trained on groundtruth sketch-3D model pairs. Then, the updater network uses the output of the first network from a random view and compares its own construction. Note that their method is aimed at professionals who are experienced in perspective drawing, while ours is for amateurs too. Similar to \cite{Delanoy2018cgit}, Sketch2CAD\cite{Li2020tog} presents a data-driven modeling system aimed at users experienced in sketching and product design, but inexperienced in 3D modeling. The authors draw parallels between the steps in a CAD modeling session and those an industrial designer follows when sketching in 2D. Motivated by these similarities, they create a tool where the user sketches the shape edits incrementally. The system automatically processes each increment into an appropriate CAD operation. 
Overall, their tool could be attractive to product designers. 

\noindent\textbf{Sketch-Based Articulated 3D Figure Modeling:} Some earlier work in sketch-based articulated figure modeling focused on stick figures. Davis~\etal~\cite{Davis2003siggraph} provide a medium for artists to create 3D animations from a sequence of stick figure sketches, with user annotated skeletal keypoints. The system is not fully automated, but gives the artist a choice to select among possible 3D poses. In addition to 3D pose lifting, \cite{Mao2006sigchi} infers the sketched character's body proportions 
and transfers it to a morphable 3D model.

Motion Doodles~\cite{Thorne2004tog} explores the task of sketching motion. The user can author a jump or somersault by drawing a path. The system supports both 2D and 3D animation for sketched characters. \cite{Jain2009siggraph} infers the 3D motion from hand-drawn sketch animations, but requires the labeling of body landmarks on the sketched body. In contrast to these methods, Akman~\etal~\cite{Akman2020arxiv} propose a deep learning approach to directly predict 3D point clouds from 2D stick figures. By interpolating the latent features of two sketches, the reconstructed 3D point clouds can be post-processed into articulated mesh models. However, their user interface does not allow for inputting start and end sketches or correcting erroneous reconstructions. 

Apart from stick figures, methods for modeling and animating more complex shapes were proposed~\cite{Levi2013cgf, Bessmeltsev2015tog, Bessmeltsev2016tog}. To accurately lift the 3D pose from input sketches, these methods require the artist to explicitly define a 3D skeleton. ArtiSketch~\cite{Levi2013cgf} requires the user to create multi-view sketches of the character. 
Along with multiple sketches, the artist must also provide 3D skeletal pose models for each view, using an external 3D tool. In \cite{Bessmeltsev2015tog}, 3D articulated figure modeling can be done from single contour drawings of cartoon characters and corresponding 3D skeletons. \cite{Bessmeltsev2015tog}'s need for explicit pose information is alleviated in Gesture3D~\cite{Bessmeltsev2016tog} but the system still requires a template 3D mesh instead of a predefined posed skeletal structure for each input sketch. The recent MonsterMash~\cite{Dvoroznak2020tog} achieves great modeling successes using only sketches, and is therefore one of our baselines. The need for a model template or 3D pose information is eliminated by the user separately annotating meaningful parts of the sketch and indicating if a part is positioned in front of neighboring parts.
In contrast to these methods, RigMesh~\cite{Borosan2012tog} combines the rigging and animation steps by automatically constructing the skeleton from contour sketches. As a variant of Teddy, which uses the Chordal Axis Transform~\cite{Prasad1997cnls} to inflate the contour sketch, RigMesh creates the skeleton of the 3D figure from the chordal axis. However, inferring the skeleton from simple contours suffers from unnatural and ill-positioned joints~\cite{Bessmeltsev2015tog}.

\noindent\textbf{3D Human Pose and Shape Estimation:} There is a rich history of human 3D pose and shape estimation algorithms for a single RGB image. Recently, classical methods \cite{Sigal2007nips, Guan2009iccv, Balan2007cvpr} have been replaced by deep learning based approaches, \eg SPIN\cite{Kolotouros2019iccv}, HMR\cite{Kanazawa2018cvpr},  \cite{Lin2021cvpr}, and \cite{Omran20183dv}. Please see \cite{Wang2021cviu} for a more in-depth survey in this field. For this work, we employ the architecture in \cite{Sengupta2020bmvc} for estimating 3D parameters from human sketches.

\vspace{-\baselineskip}
\section{Method}
Our system enables a user with a digital stylus to quickly position the limbs of an articulated 3D human ``mannequin,'' mostly by means of figure sketching, as illustrated in Fig~\ref{fig:teaser}.  An overview of our method is in Fig~\ref{fig:network}. In Sec~\ref{subsec:figure_sketches}, we describe our underlying generalized-cylinder-like representation~\cite{marr1978representation} called the 3D Primitive Human Body Model. To overcome a lack of real and varied training data for sketch-based 3D pose estimation, we introduce our core synthetic data augmentation strategy for generalizing this model to artist drawn sketches in Sec~\ref{subsec:sketch_aug}. In Sec~\ref{subsec:sketch_to_3d} we describe the model we use for predicting human pose from 2D sketches. 

\subsection{Figure Sketching} \label{subsec:figure_sketches}
\begin{figure}[ht]
\vspace{-\baselineskip}
  \includegraphics[width=.45\textwidth]{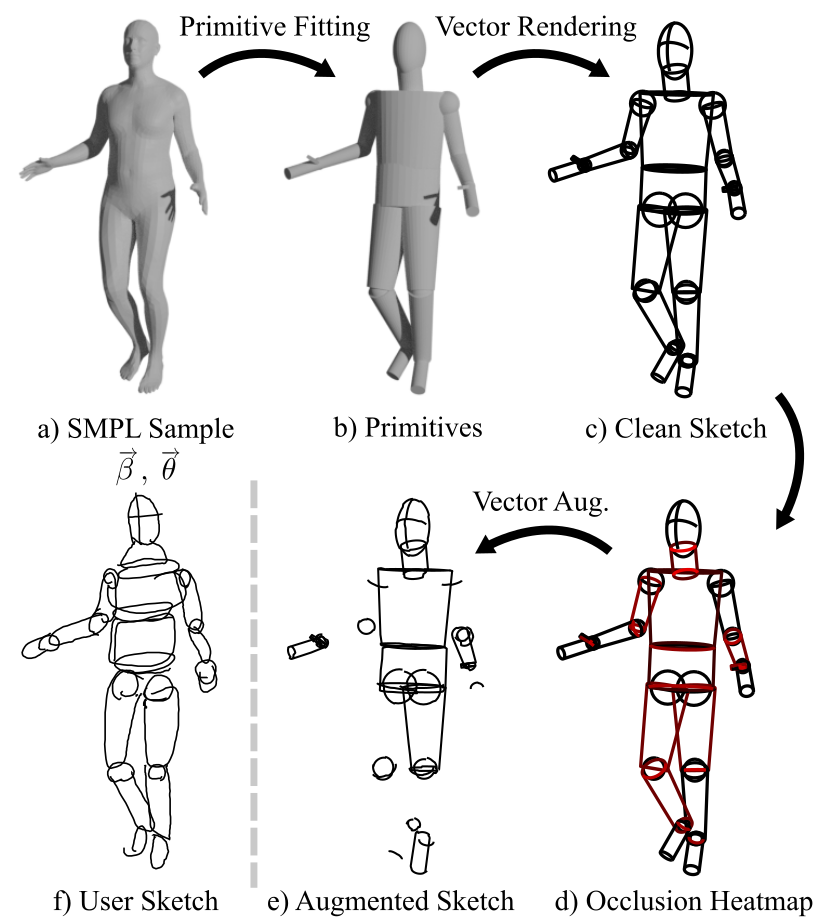}
  \vspace{-.8\baselineskip}
  \caption{Overview of our synthetic sketch generation. a) we generate a sample SMPL body with pose ${\ora{\theta}}$ and shape ${\ora{\beta}}$. In b) we fit geometric primitives to the generated body. We then render these primitives down to clean vector sketch lines in c). d) While all joints and vector strokes are susceptible to being deleted as part of augmentation, strokes with more hidden nodes are given a higher probability of being deleted. e) Our augmentations include local node translational jitter, deleting strokes and joints, and global stroke translations. f) An example \emph{user} sketch from our user study.}
  \label{fig:sketch_generation}
  \vspace{-\baselineskip}
\end{figure}

For human figure sketching, artists commonly use either their imagination or reference images as inspiration. To quickly convey the reference pose of a human, regardless of the details in the inspiration, one method is to use simple primitive shapes for body parts, as advocated by Lee~\&~Buscema~\cite{Lee1984marvel}. Inspired by this, we created the \emph{3D Primitive Human Body Model} (\textbf{3DPHB}) in Fig~\ref{fig:sketch_generation}.b. 

3DPHB has correspondences to the SMPL parametric body model~\cite{Loper2015tog}. Figs.~\ref{fig:sketch_generation}a-b show an example of how our Primitive Human relates to SMPL bodies: we sample a set of body shape and pose parameters (${\overrightarrow{\beta}}$, ${\overrightarrow{\theta}}$) and place the following primitives:
\begin{itemize}
    \itemsep0em 
    \item a 3D ellipsoid for the head,
    \item tapered cylinders with varying radii at each base for limbs,
    \item spheres for joints, 
    \item and two tapered cylinders for upper and lower torso.
\end{itemize}
Note that we draw samples with very distinct poses, but with little variation in the shape, so the generated mannequins will differ mostly just in pose. Informally, we observed that allowing shape to vary significantly would mean that beginner artists got less predictable results. The lengths of body part primitives are directly aligned with the original body. The upper torso and lower torso are aligned to the width of the shoulders-waist and waist-hip segments, respectively. The head is sized w.r.t. the top of the head \vs neck, and to fit between the ears. 

\subsection{Part-Aware Augmentations for Figure Sketches}\label{subsec:sketch_aug}

A major obstacle for training CNNs for sketch-based tasks is the lack of sketch datasets with corresponding 3D labels. Paired data for sketch-based 3D reconstruction would ideally span various poses and drawing styles, though we view ``style'' as mostly meaning artist thoroughness and precision. Thus, for 3D pose prediction from sketches, we generate a synthetic sketch and 3D human pose dataset coupled with a vector graphics augmentation scheme to generalize to human-made sketches. 

Using our 3DPHB model, we produce sketch-like renderings using a set of different line types: silhouettes, contours, creases, and borders, as can be seen in Fig.~\ref{fig:sketch_generation}.c. To convey the head orientation, a vertical and horizontal line are used for the eyes and nose. 

Renderings are stored into a vector graphics format for our synthetically generated sketches. This format is extremely flexible and useful for storing extra information such as per-stroke body part labels and occlusion. We associate each stroke with a corresponding body part label. Such labeling of the 2D renderings allows us to control the types of augmentations we can apply to each body part during CNN training. More specifically, for a given camera pose, we use ray casting operations to determine whether a stroke is visible or occluded. Since each body part could consist of several strokes, we assign each part with an average occlusion rating based on this information. Our body part-aware sketch augmentations include global translation of body parts, local stroke jitter, part-based hiding tied to occlusions, and random part-based hiding. The latter two are needed to simulate two artist behaviors we observed in a pilot study. First, people differ in which primitive lines they decide to put on the canvas. Depending on preference, the artist could decide to keep or discard strokes for occluded body parts. Also, people tend to forget to draw the sketch lines for some body parts, especially at already-crowded joints.


More formally, our part-aware figure sketch augmentations can be defined as a set of transformations applied to each body part. We denote an undisturbed figure sketch rendering as set $S$ and an augmented sketch as $S^{*}$. We denote all augmentation operations as $aug(*, A)$ where $A$ is the set of part augmentations that affect different body parts. $S$ and $S^{*}$ have the relationship that $S^{*} = aug(S, A)$, where $S$ consists of a set of body parts $B=\{j, l, t, h\}$. Here, $j$, $l$, $t$, and $h$ denote joints, limbs, the two-piece torso, and the head with neck. Each body part $b\in B$ could consist of several strokes $s_{b}$ and different augmentations could affect either a complete body part or a single stroke. Specifically, we use the following set of augmentations: $A=\{ translate, jitter, hide_{random}, hide_{occluded} \}$. A body part could be translated by sampling translation offsets $b^* = b + t$, where $t \sim \mathcal{N}(\mu, {\sigma}^2)$ is a randomly sampled from a Gaussian Distribution with mean $\mu$ and variance $\sigma^2$. Similarly, we add jitter to strokes, so $s_{b}^* = s_{b} + t, t \sim \mathcal{N}(\mu, \sigma^2)$. Another type of augmentation is hiding of body parts. We simulate missing body parts by turning off sketch lines randomly. Further, to implement occlusion-based hiding, we give an occlusion rating to each stroke: given a stroke $s$ consisting of subnodes $n_{s}= \{n_1, n_2, n_3, ...,  n_n\}$ with total number of occluded nodes $v \leq n$, the occlusion rating $o_s = \frac{v}{n}$. Thus, a high $o_s$ implies a higher probability of the stroke $s$ being hidden.

\subsection{3D Pose Estimation from Sketches} \label{subsec:sketch_to_3d}
To predict the 3D body pose from a sketch, we use a two-step pipeline, depicted in Fig.~\ref{fig:network}. Given an input figure sketch, our Sketch Interpreter 
infers 2D joint locations and silhouettes. We use these intermediate representations between sketch input and 3D output due to the nature of the sketch medium in general~\cite{Yue20203dv}. Inferring 3D information from 2D input alone is ambiguous. Sketches are sparse in nature and compared to RGB images, they contain fewer 3D cues, lacking textures and shadows. Those cues are available for 3D lifting tasks that start with real photographs. Previous work in both image-based\cite{Thai2020arxiv} 3D reconstruction shows that using silhouettes as an intermediate representation does help. 

The 3D lifting stage is used to predict 3D pose information using the intermediate representations inferred from the input sketches. As a means of lifting, we incorporate the sketching-unaware pre-trained 3D network from STRAPS~\cite{Sengupta2020bmvc}, which takes silhouettes and 2D joint locations as input, to infer the 3D shape and pose parameters that we seek, to reflect 
the upstream input sketch.

\begin{figure*}[ht]
\centering
\includegraphics[width=0.8\textwidth]{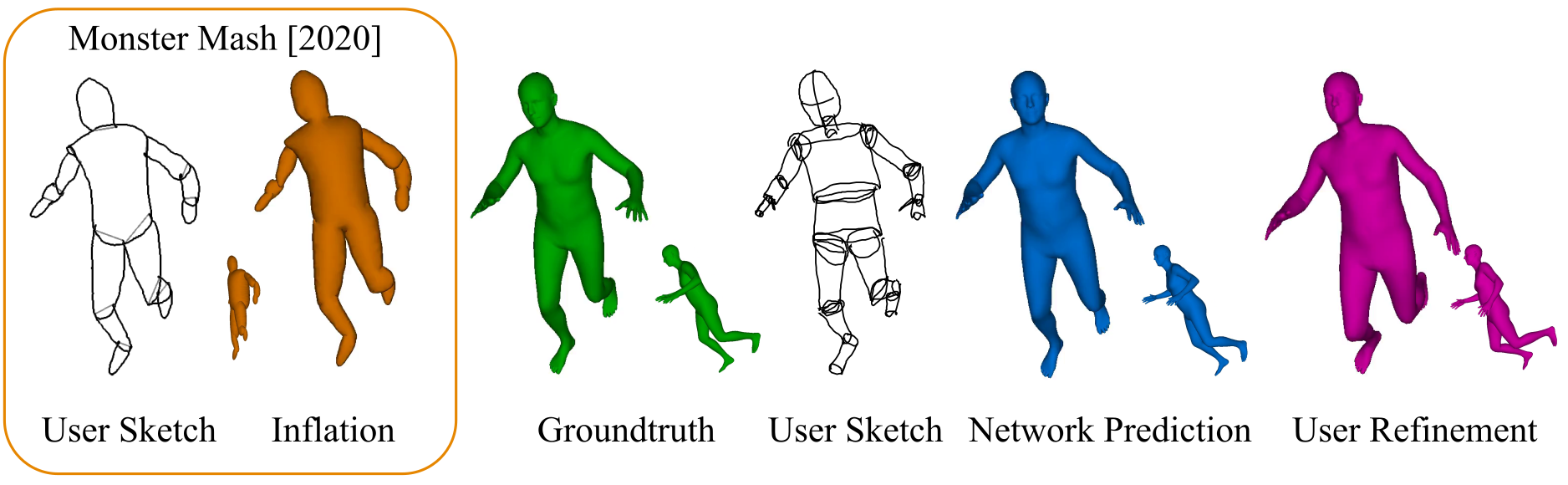}
 \vspace{-\baselineskip}
  \caption{Example sketch, intermediate output, and refinement from our user study and MonsterMash experiment. For each model, we show both a front view (large) and a side view (small). The green model shows the frontal pose that was given to users of both systems as a reference. Highlighted with orange is a user's sketch input to MonsterMash~\cite{Dvoroznak2020tog} and the resulting 3D shape. Note that the grey lines in the MonsterMash sketch are auto-completed as an intermediate step in their pipeline. Our initial network prediction (blue) is already faithful to the input sketch, and the groundtruth (green); the user chose to refine further via classic 3D controls (purple). MonsterMash does more than posing, but since its inflated models start in a planar world, much of the information on depth embedded in the sketch is lost.}
   \vspace{-\baselineskip}
  \label{fig:monstermash}
\end{figure*}

\emph{\textbf{Implementation Details}} Our Sketch Interpreter is built on top of the codebases of DensePose~\cite{Guler2018cvpr} and Keypoint-RCNN~\cite{He2017iccv}. Both networks are trained on our synthetic sketch dataset from scratch. DensePose predicts a mapping between images of humans and the surface of a template 3D model, and is normally trained using a manually annotated dataset of humans. For our task, we generate a synthetic dataset with dense surface correspondences between synthetic sketches and the 3D body template, along with body part segmentation maps and 2D joint locations. There are 25K sketches in our dataset. We train all models using \cite{Guler2018cvpr} and \cite{He2017iccv}'s default hyperparameters for 100K epochs with a learning rate of 0.002 on a single NVIDIA Titan Xp 12GB. We lift 2D joints and silhouettes using a pretrained STRAPS~\cite{Sengupta2020bmvc} network. All models were implemented using PyTorch~\cite{pytorch}. The code will be made ready on GitHub.

\section{Experiments}

\begin{figure*}[t]
\centering
\includegraphics[width=0.91\textwidth]{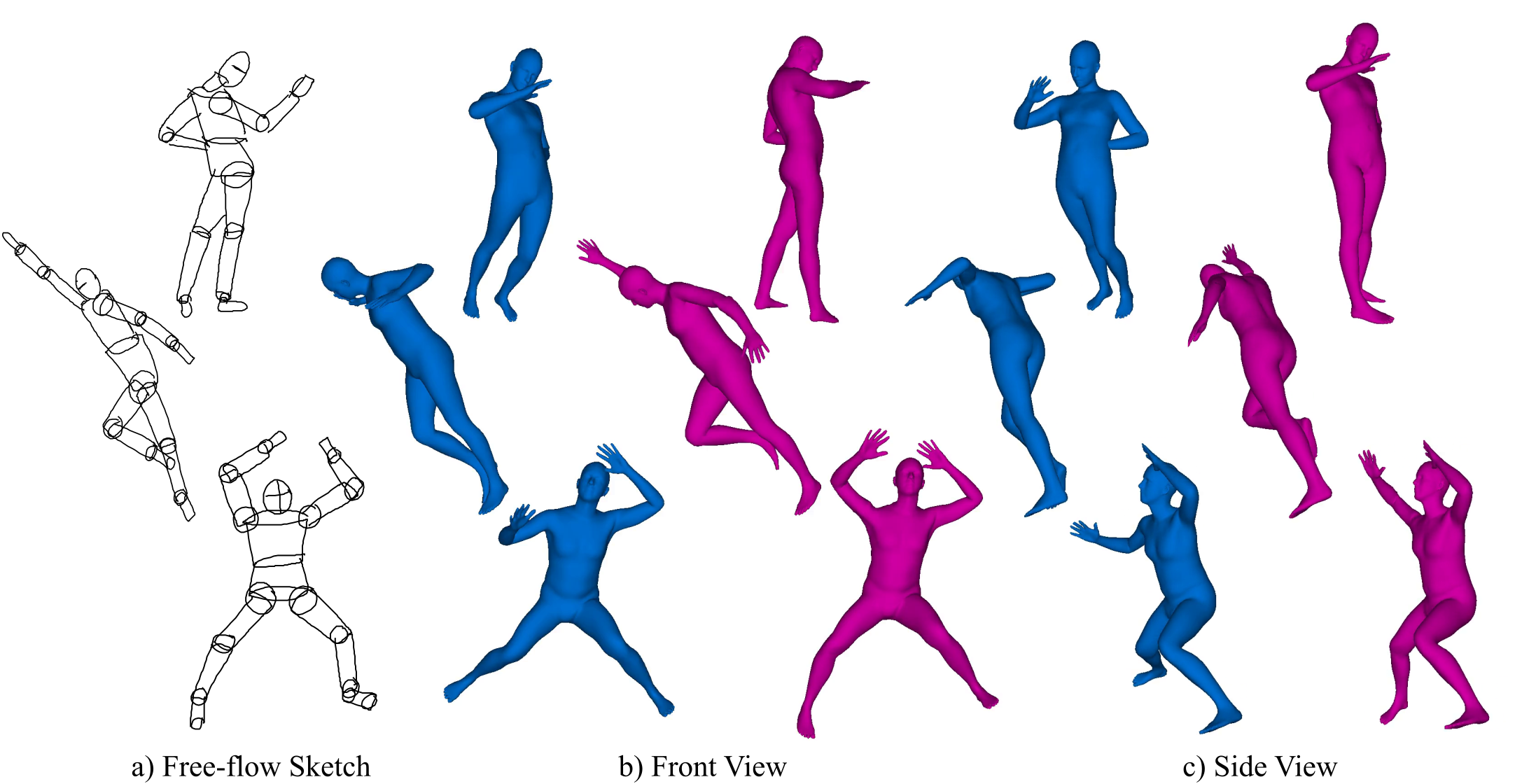}
\vspace{-1\baselineskip}
  \caption{Some users (separate from the evaluated user-study) were allowed to sketch on a blank canvas. Some sketches are in a). Users (one for each row) then refined the network's predictions (blue) to get the pose that they were really after in 3D (highlighted in purple), potentially deviating from the pose they had in mind initially.}
  \vspace{-\baselineskip}
  \label{fig:cool_results}
\end{figure*}

We validate our system and its components in three ways. First, a user study in Sec.~\ref{subsec:user_study} compares our approach in the context of posing 3D mannequins to match a reference image. The evaluation metric is in Sec.~\ref{subsec:numbers_eval}. Second, we perform ablations of our sketch generation pipeline, and comparisons to ~\cite{brodt2022sketch2pose} and against a state-of-the-art sketch augmentation baseline~\cite{wailly2019line} in Sec.~\ref{subsec:ablations_competition}. Third, we demonstrate qualitative results of \emph{interactive} user sketching and mannequin posing here and in the supplemental video.

\subsection{User Study}\label{subsec:user_study}
We evaluated our system with a collection of novice users. This means they had a variety of backgrounds, but had never done sketch-based modeling or similar tasks. We define two modes, so users were asked to either 1) adjust a human body from a canonical pose (T-pose) using 3D handles, or 2) users were asked to first sketch out a human pose, allowing our model to predict a mannequin pose to replace the T-pose, and then continuing to refine the pose as in the first mode. To help even the playing field between amateur artists and to make the results more measurable, users were given a reference render of a known ground truth posed human mesh that they must aim to closely mimic. The reference renders were generated using the same 3D body shape that users refined in both modes.

\textbf{User Interface} We built our UI to allow for both 2D and 3D manipulation of sketches, and direct manipulation of a 3D posed human. Users can draw strokes using their fingers or a stylus on a touch screen, or a mouse, though all our users opted to use a stylus for their sketches. We used a combination of three.js, Blender, JavaScript, and our Python backend. Users were allowed access to the UI via a browser, making it readily accessible on most consumer electronic devices. For 3D refinement of rigged meshes, we adapted the Rigify extension in Blender to post-process network predictions. This allowed us to add joint limits and expose interactive 3D models amenable to control via Forward Kinematics (FK) and Inverse Kinematics (IK). These controls are available to the user in both study modes.

\textbf{Study Details} We invited 12 users to participate in our user study. We asked users to go through a four-step process: a tutorial video, a practice period, and both a session in the manual T-pose refinement mode and a sketch-to-3D-session. The tutorial video introduced users to our UI and gave them an overview of the generic task they needed to accomplish along with illustrations of how to sketch figures and how to use FK and IK. The practice period was 10 minutes long and allowed users to become familiar with the UI, figure sketching, and 3D controls. After the practice session, users were asked to finish both tasks sequentially, with 10 minutes for each. We randomized the order of tasks and the reference render for each user. We had arbitrarily chosen the parameters controlling the amount of data augmentation in the training of our model. Later, we will refer to this parameter setting as Augmentation 2. 

\textbf{Data Collected} All user actions, including sketch strokes and timing, were recorded for evaluation. To measure how much each user spent for each task, we compute how much time was spent on sketching and separately on using 3D FK/IK controls to the point where the user was satisfied with the result. We also collected feedback from users via a questionnaire at the end of the user study. 

\subsection{Evaluation Pipeline}\label{subsec:numbers_eval}

We perform evaluation using two 3D metrics: Chamfer Distance on sampled point clouds and Mean Per Joint 3D Position Error (MPJPE)~\cite{Mehta20173dv} on the underlying joints. For both modes of the user study, we report metrics comparing the refined 3D models to the ground truth, averaged across all user sessions. For Sketch Refinement, we also evaluate the initial prediction from our system from the initial user sketch, before any further refinement. We've found that users may often misjudge the orientation of meshes along depth given the reference 2D image. To account for this, all 3D metrics are computed after meshes have been aligned, as a rigid body, to the groundtruth using Iterative Closest Point (ICP)~\cite{besl1992method}.

\subsection{Ablation Study}\label{subsec:ablations_competition}
We subsequently ran an ablation study to quantify the effect of our sketch augmentations on sketch-based 3D pose estimation. We trained our Sketch Interpreter (Fig~\ref{fig:network}) with three levels of sketch augmentation severity, so based on Ours(Default) plus and minus 10\%. More specifically, we experiment with the severity of our part-aware augmentation strategy. We train two more baseline models without vector-based augmentations: our sketches and our sketches but with a single-piece torso in-place of two.  We also compare to the network and optimization of Sketch2Pose\cite{brodt2022sketch2pose}.

\subsection{Baseline Systems}
\begin{figure}[ht]
\centering
\includegraphics[width=0.43\textwidth]{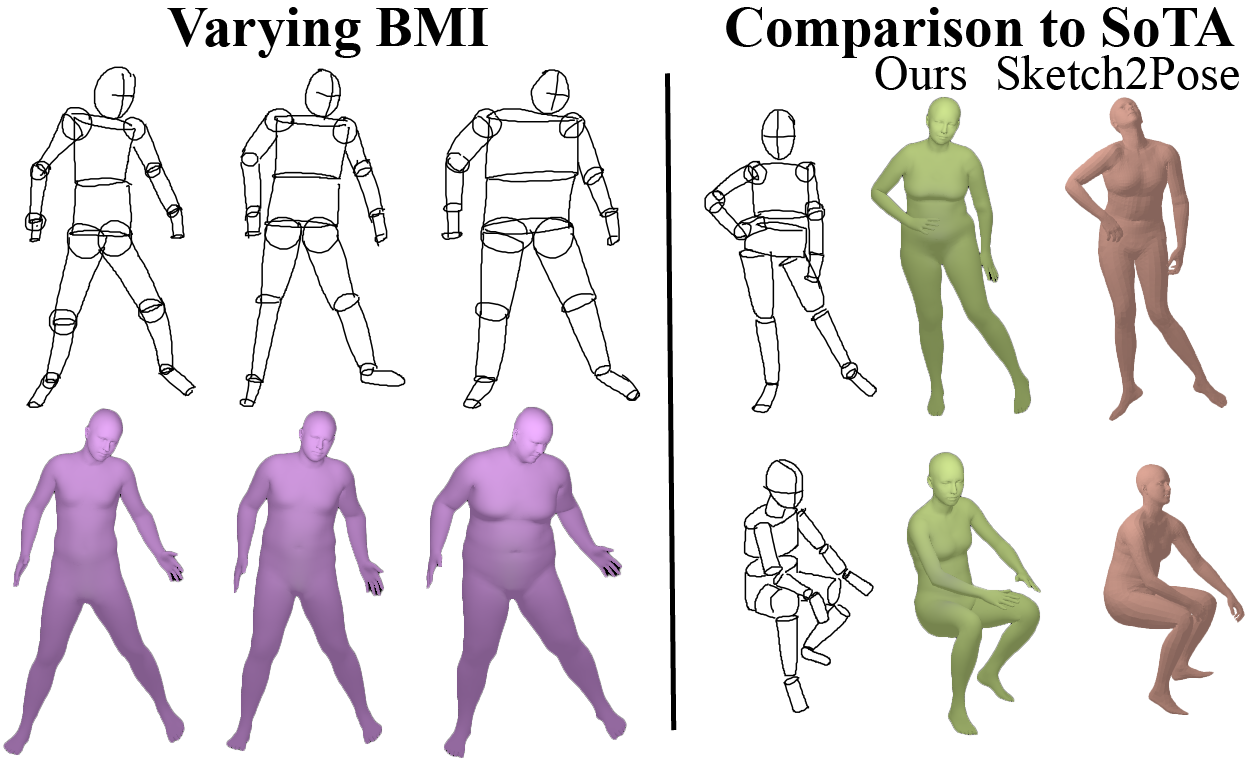}
\vspace{-1\baselineskip}
  \caption{(Left) Like Sketch2Pose\cite{brodt2022sketch2pose}, we normally fix shape, and infer only pose. Here exceptionally, are our pose \emph{and shape} estimation results for sketches with varying BMI. (Right) Generalizing to unusual poses may prove challenging, but our current results are on par with Sketch2Pose, pictured. Also, ours is 560x faster and allows the artist to refine the result further.}
  \label{fig:bmi}
  \vspace{-1.4\baselineskip}
\end{figure}

Sketch2Pose~\cite{brodt2022sketch2pose} was developed in parallel to our approach, and represents the current SoTA in pose-from-sketch inference. We compare against it in Table~\ref{tab:ablation_results}, though the comparison is somewhat unfair to us, evaluating their offline computer vision system to our interactive real-time system for artists to sketch and iterate. Their supervised training lets them interpret sketches as 2D joint locations, but then requires a costly 90 sec. optimization to produce a mesh. 
Our method performs comparably to Sketch2Pose on real sketches, in a fraction of the time, as seen in Fig.~\ref{fig:bmi}~(Right). We compare against this one-shot image-to-model converter because of the partial overlap in their initial stage, despite ours being a real-time system without reliance on labeled sketches. 

To evaluate our body part and occlusion aware vector augmentations, we train our Interpreter with the vector augmentation baseline scheme from SynDraw~\cite{wailly2019line}.

MonsterMash~\cite{Dvoroznak2020tog} requires a very specific set of input strokes for sketches to be valid, so our user study sketches do not produce meaningful results in their tool. We do however compare qualitatively on an adapted reference sketch in Fig.~\ref{fig:monstermash}. With 30 minutes of practice, a user was able to sketch meaningful input that would produce the best available inflated mesh. This comparison too is somewhat unfair, because the authors of that system could likely teach our users more about the intended drawing rules.

\section{Results}
We present qualitative and quantitative results for our interactive sketch-based 
system. All quantitative results are reported on real sketches collected during our user study, using 
\textit{Ours(Default)} settings, as shown in Tab~\ref{tab:ablation_results}. 

\textbf{User Study} We report quantitative results from our user study in Table~\ref{tab:user_study_results}. Sketch+Refine performs similarly to classic manual refinement on chamfer distance while achieving better scores on 3D joints. On average, sketching alone was almost four times as fast as manual refinement and achieves competitive accuracy. When adding the extra time needed by some users to refine our prediction, the total time remains shorter compared to manual refinement. Crucially 70\% of our users preferred using sketching and partial refinement to just manual refinement. 

\textbf{Ablation} We ablate our system in Table~\ref{tab:ablation_results}. We use all 48 real human-drawn sketches from our user study (includes practice sessions) for comparing our ablated model and competitors. All three of our augmentation schemes perform better than or similarly to baseline methods. Among those, Ours(Heavy), which is the scheme with the most severe augmentations, outperformed all baselines and competitors. We observed that users tend to forget to put down strokes for body parts; thus, training with missing strokes and body parts better helped to generalize to real sketches. 

\textbf{Qualitative Results} Please refer to Figure~\ref{fig:cool_results} for free-flow sketches and to Figure~\ref{fig:monstermash} for a sample of a user study sketch. Please also see the supplemental material. We limit the scope of this paper to human pose estimation from sketches. However, our system can interpret simple shape variations, as illustrated in Fig.~\ref{fig:bmi}~(Left).
  
\begin{table}[ht]
 \begin{center}
\begin{tabular}{|l|ccc|}
            \hline 
                    & Chamfer$\downarrow$ &  Joint3D$\downarrow$  & Time$\downarrow$\\ 
            \hline  \hline 
            Canonical Pose &          0.02931   &         0.2647 &                -   \\
            Manual Refine.  &         0.00730   &         0.1208 &             402.35s\\
            Sketch Prediction  &      0.01006   &         0.1224 &     \underline{135.498s}\\
            Sketch + Refine. &\textbf{0.00652}  & \textbf{0.0933} & \textbf{313.877s}\\
            \hline 
          \end{tabular}
 \end{center}
  \vspace{-1.2\baselineskip}
 \caption{Quantitative results of the user study with 18 participants. Distances are in meters. Our prediction from user sketches alone scores competitively compared to manual refinement, while taking a quarter of the time on average. Further interactive manual refinement (Sketch + Refine) improves mesh and joint metrics, while still taking little time.}
 \label{tab:user_study_results}
 \vspace{-1\baselineskip}
\end{table}

\noindent
\begin{table}[ht]
\center
\begin{tabular}{|l@{\hspace{.45\tabcolsep}}|c@{\hspace{.45\tabcolsep}}c@{\hspace{.45\tabcolsep}}c@{\hspace{.45\tabcolsep}}c@{\hspace{.45\tabcolsep}}|}
         \hline   
          & Chamf.$\downarrow$ &  Joint3D$\downarrow$  & Joint2D$\downarrow$ & MPVPE$\downarrow$ \\ 
        \hline  \hline  
        Sing.Torso  &      0.02097 & 0.2595 & 69.47 & 0.3126\\ 
        Doub.Torso  &      0.0143 & 0.1825 & 53.08 & 0.2260\\ 
         \hline  
        SynDraw~\cite{wailly2019line}  &   0.0245 & 0.2537 & 92.65 & 0.3007\\
        Ske2Pose~\cite{brodt2022sketch2pose}  &   0.0070 & 0.1682 & 21.09 & 0.1607\\
        Ours(Def.) &   0.0086 & 0.1149 & 19.54 & 0.1391\\  
         \hline \hline
        Ours(Light) &   \textbf{0.0065} & \underline{0.1031} & \underline{18.62} & \underline{0.1300}\\ 
        Ours(Heavy) &   \underline{0.0069} & \textbf{0.0953} & \textbf{18.20} & \textbf{0.1230}\\
       \hline
       \end{tabular}
       \vspace{-0.6\baselineskip}
       \captionof{table}{Quantitative comparison of our augmentation method and its ablations against baselines. We used Default augmentation (so untuned for this test set) for our user study, which is already competitive against \cite{brodt2022sketch2pose} (which requires an extra 90 second optimization). We also show tests of our model with $\pm$10\% augmentation, which indicates novices could benefit from heavier augmentation.}      
      \label{tab:ablation_results}
\end{table}
\vspace{-\baselineskip}
\section{Conclusion}
We demonstrated a highly interactive system that allows users to sketch the desired 3D pose of their mannequin, in ``The Marvel Way''~\cite{Lee1984marvel}. The system provides multiple methods for refining the estimated pose, which is important to users for the creative process to be more than just a curiosity. Surprisingly, manipulating even IK handles is slow and cumbersome enough, that users starting from a T-Pose had a disadvantage compared to starting by sketching out the body using primitives. Sketching required very little training and was also reported to be more fun!
\begin{figure}[t]
\centering
\vspace{-1\baselineskip}
\includegraphics[width=.4\textwidth]{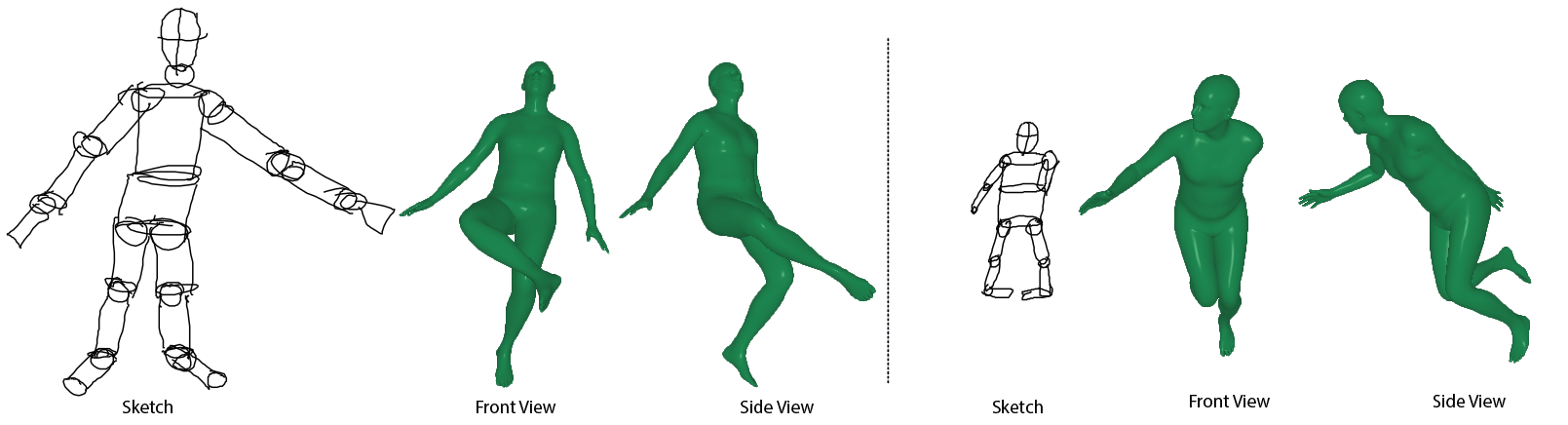}
\vspace{-1\baselineskip}
  \caption{Limitations: our system can produce unexpected mannequin poses when provided with limb lengths that are implausible (left) or out-of-distribution sketches, \eg input that is too small.}
  \vspace{-1.4\baselineskip}
  \label{fig:limitations}
\end{figure}

The system has limitations (shown in Fig~\ref{fig:limitations}) that are obvious once the artist draws body shapes that are very different from the training data. This is a problem for very unusual poses, such as drawing upside-down people. Future work will explore the trade-offs when training a model for a variety of poses and body shapes. Until then, the resulting SMPL-based mannequin can be manipulated using sliders (not in our UI) to achieve other shapes. Scaling to other creatures is achievable using our synthetic vector-graphics renderer.

\noindent\emph{Acknowledgements} We would like to thank Prof. Iasonas Kokkinos for valuable guidance and PhD funding from Niantic and Microsoft.

{\small
\bibliographystyle{ieee_fullname}
\bibliography{egbib}
}

\end{document}